\documentclass[sigconf]{acmart}

\AtBeginDocument{%
  }

\setcopyright{rightsretained}
\acmConference[SIGIR eCom'22]{ACM SIGIR Workshop on eCommerce}{July 15, 2022}{ Madrid, Spain}
\acmYear{2022}
\copyrightyear{2022}
\makeatletter
\renewcommand\@formatdoi[1]{\ignorespaces}
\makeatother
\acmISBN{}

\begin{document}

\title{Product Review Image Ranking for Fashion E-commerce}

\author{Sangeet Jaiswal}
\email{sangeet.jaiswal@myntra.com}
\affiliation{%
  \institution{Myntra Design Pvt Ltd}
  \city{Bangalore}
  \country{India}}

\author{Dhruv Patel}
\email{dhruv.patel@myntra.com}
\affiliation{%
  \institution{Myntra Design Pvt Ltd}
  \city{Bangalore}
  \country{India}}

\author{Sreekanth Vempati}
\email{sreekanth.vempati@myntra.com}
\affiliation{%
  \institution{Myntra Design Pvt Ltd}
  \city{Bangalore}
  \country{India}}
  
\author{Konduru Saiswaroop}
\email{konduru.saiswaroop@myntra.com}
\affiliation{%
  \institution{Myntra Design Pvt Ltd}
  \city{Bangalore}
  \country{India}}

\renewcommand{\shortauthors}{Jaiswal et al.}

\begin{abstract}
In a fashion e-commerce platform where customers can't physically examine the products on their own, being able to see 
other customers' text and image reviews of the product is critical while making purchase decisions. Given the high reliance on these reviews, over the years we have observed customers proactively sharing their reviews.
With an increase in the coverage of User Generated Content (UGC), there has been a corresponding increase in the number of customer images. It is thus imperative to display the most relevant images on top as it may influence users' online shopping choices and behavior. In this paper, we propose a simple yet effective training procedure for ranking customer images. 

We created a dataset consisting of Myntra (A Major Indian Fashion e-commerce company) studio posts and highly engaged (upvotes/downvotes) UGC images as our starting point and used selected distortion techniques on the images of the above dataset to bring their quality to par with those of bad UGC images. We train our network to rank bad-quality images lower than high-quality ones. Our proposed method outperforms the baseline models on two metrics, namely correlation coefficient, and accuracy, by substantial margins.

\end{abstract}



\maketitle

\section{Introduction}
Online purchases have seen tremendous growth over the last few years \cite{statista}. In 2021, there were approximately 190 million customers annually, compared to 135 million customers in 2019 in India. This increase can be attributed to the growth of the e-commerce industry and to the COVID-19 pandemic, which led to a change in the shopping behavior of customers. A similar trend is expected in 2022 too. 
    
Accurate, non-misleading visual representation is essential for fashion products sold via e-commerce platforms. Catalog images, at Myntra, are taken under well-exposed environments. However, users tend to trust images generated by other users more than those generated by the brands\cite{ugcsurvey}. At Myntra, we allow our users to optionally post their photos along with the review they write. However, user-generated images sometimes, do not comply with the required standards of accuracy of visual representation. For instance, some photos are mirror selfies taken at awkward angles, some are under-exposed, while others are over-exposed, and many are cropped representations of the products. In this paper, we propose a Machine Learning Model that can differentiate ``good'' UGC images from ``bad'' UGC images. With the help of this model, we can sort available UGC images by their quality so that the users do not have to browse too many images to get the feel of the product they are examining.

How can we tell if the image is good or bad? In an ideal world, we can either explicitly ask users to rate the images on some scale, and based on the collective wisdom, sort the available images, or we can implicitly log the average time spent by the users on each image, and then sort the images that are viewed for longer duration first. The explicit approach is not feasible as we have to decouple UGC images from the UGC reviews and ask users to rate both, making the experience inconvenient. The implicit approach requires us to augment the current infrastructure to log the average time spent metric.

An alternative approach would be to sort the images based on the "likes to total ratings ratio" based on the votes on associated text reviews. This approach does not work well when we have new reviews or relatively lesser-explored products. At Myntra, 50\% of the reviews containing at least one photo does not have any votes at all, while 75\% of such reviews have less than five votes. One can use bandit-based approaches\cite{slivkins2019introduction} to solve this cold start, but those too will require some augmentation to the existing system to capture how users are reacting to the ranking.

Our model is trained on the image pairs generated in such a way that one image will almost always be superior to another one in quality. We train a multi-layer perceptron to score good images higher than bad ones using pairwise hinge loss. To generate such a dataset, we make certain assumptions. One such assumption is that a professionally taken image will be better than a user-generated image. Another assumption is that for highly engaged reviews, users also consider the quality of the associated image while deciding whether that review is helpful or not. 

Our key contributions are,
\begin{enumerate}
    \item We propose an effective learning scheme by leveraging pre-trained models to extract features for image aesthetic assessment in fashion e-commerce without manual annotation. 
    \item To the best of our knowledge, this is the first attempt at ranking fashion UGC images.
\end{enumerate}

The rest of the paper is organized in the following way. Section 2 gives an overview of the related work. Section 3 explains how we generate a synthetic ranked dataset and our approach to learning the ranking. In Section 4, we explain the experimental setup. Results are given in Section 5. Section 6 concludes the work.

\section{Related Literature}

The recent trends in approaching the Image Aesthetics Assessment (IAA) problem have been based on either regression or classification. Most of these models use the AVA\cite{murray2012ava} or AADB\cite{kong2016aesthetics} datasets to benchmark their performance.

Yeqing et al.\cite{wang2016finetuning} consider the IAA as a binary classification task where they segregate images based on their Mean Opinion Score (MOS). Images with MOS less than 5 will be treated as bad images, and those whose MOS is greater than or equal to 5 are considered as good images. They finetune Convolution Neural Network (CNN) models pre-trained on ImageNet such as AlexNet and VggNet to report their accuracies. There are other approaches that deal with the problem of fixed-size image constraints of CNN \cite{ma2017lamp,he2015spatial,mai2016composition,wang2020image} but eventually, they also solve IAA as binary classification.

Neural Image Assessment (NIMA) \cite{talebi2018nima} introduced a simple strategy. While most of the then-existing approaches were based on predicting the MOS, they predicted the aesthetic rating distribution using a CNN that is trained using Earth Mover’s Distance Loss (EMD) on human-sourced rating distribution from AVA dataset. Despite its simple architecture, it achieves a result that is comparable to State of the art results. We have adopted NIMA for our experiments. We have used MobileNet\cite{howard2017mobilenets} architecture-based CNN as our backbone network to generate image features that are trained on AVA and TID2013\cite{ponomarenko2015image} datasets.

A task related to IAA is No Reference - Image Quality Assessment (NR-IQA) which assesses the technical quality of an image. Many recent approaches\cite{bosse2017deep,kang2014convolutional,talebi2018nima} make use of labeled data such as TID2013, LIVE\cite{sheikh2006statistical} and CSIQ\cite{CSIQ} to predict the quality score. Another set of approaches treats this task as a ranking problem and tries to minimize the ranking loss using ground truth labels\cite{chen2009ranking,sculley2009large}. One of the drawbacks of using deep learning-based NR-IQA methods is the need of a large labeled dataset which is not available for NR-IQA tasks. The annotation process for IQA image datasets requires multiple human annotations for every image. This process of collection annotation is very time-consuming and expensive, due to which all the above approaches train shallow networks directly on the dataset. 
To address this problem,  
Liu et al. in RankIQA\cite{liu2017rankiqa} paper have taken large unlabeled high-quality images and applied image distortions to generate a ranking image dataset. For example, given an image, upon the addition of Gaussian blurs of
increasing intensities on it, we end up with a set of images which can be ranked easily as Gaussian blur will decrease the image quality. In such datasets, we don't have the absolute aesthetic score of an image, but we certainly know for a pair of images which one is of the higher quality. This synthetic data generation allowed them to better train a deep network. Subsequently, they trained a siamese network\cite{chopra2005learning} using efficient ranking loss and further fine tuned the network on a labeled dataset to achieve better performance in NR-IQA task.

Our approach is inspired by RankIQA, which distort the technical aspect of high-quality images by adding Gaussian noise, gaussian blur, etc. to generate a ranking dataset, we are using image manipulation techniques that not just degrade the technical aspect but other aspects of image qualities as well which we generally encounter in our ``bad'' UGC images.
We have also created a pair-sampling strategy that is suited to our use case of ranking images, this strategy narrows the scope of learning and provides more consistent training to our network.

\section{Methodology}

Recent IAA methods rely on training CNN that receives an image as an input and generate a score that is higher for an aesthetically superior image. These networks are generally pre-trained on the ImageNet dataset and further trained end-to-end on AVA or TID datasets for image aesthetics or technical assessment. However, such a trained network performs suboptimally in ranking domain-specific images directly because of the high diversity in the image content of these datasets.

\subsection{Data Collection}
In the RankIQA\cite{liu2017rankiqa} paper, high-quality images were subject to different kinds of image manipulation techniques with different control parameters. Applying such techniques to high-quality images ensures that introducing any type of distortion will certainly degrade the aesthetics. Using this approach as a reference point for creating a synthetic dataset, prior to introducing pertinent distortions, we started off by compiling 19k Highly aesthetic images from Myntra Studio, and to prevent the network from
overfitting to a specific type of image creation style we sampled 16k highly upvoted UGC images drawn from customer reviews. After introducing distortions to the compiled images, to the resulting dataset, we added another 3.5k highly downvoted UGC images.

\begin{figure}[t]
    \centering
    \includegraphics[width=.4\textwidth]{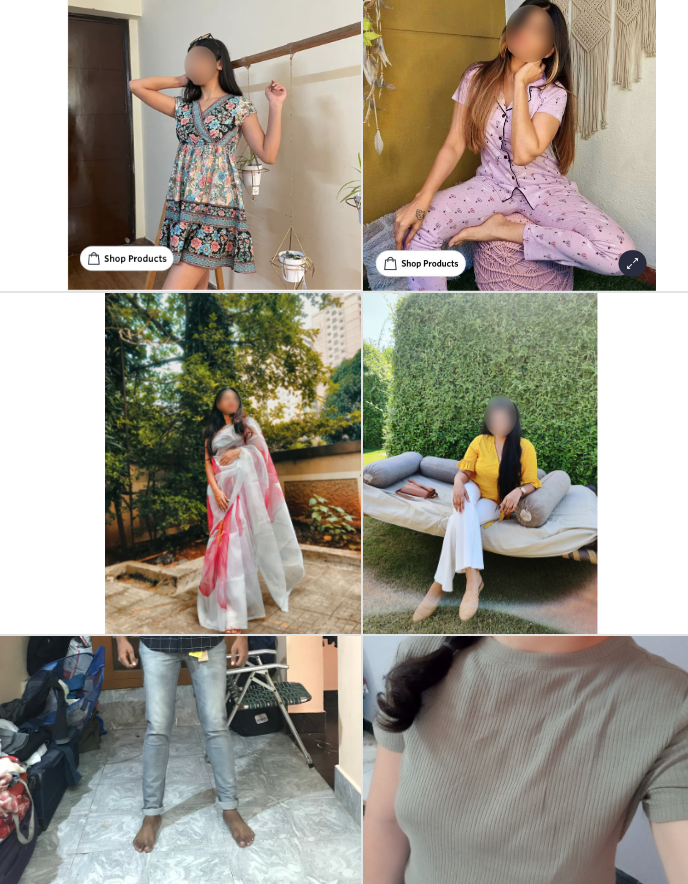}
    \caption{Sample training set images. The first row represents Studio images, second and third row represents good UGC and bad UGC images, respectively.}
    \label{fig:training_samples}
\end{figure}

\begin{description}
    \item[Myntra Studio]
    Myntra studio is a platform where fashion influencers post their images/videos wearing products that one can buy from Myntra.
    
    \item[UGC Images]
    In Myntra, a verified buyer can write reviews and can upload images to support their opinion. Any customer can upvote or downvote a review. 
    
\end{description}
A sample representative image from the training set is shown in figure \ref{fig:training_samples}.

\subsection{Image Manipulations Techniques}

As described in the papers \cite{liu2017rankiqa,sheng2020revisiting}  different image manipulation techniques have diverse effects on the manipulated image. For example, in grayscale image conversion, it is difficult to compare the input with the output in terms of aesthetics. But in our case we want to rank such images lower than the colored counterpart because it will be hard for the customer to make sense of the color of the product in a grayscale image.
Likewise, we have identified certain image manipulation techniques that are guaranteed to render a degraded effect on the image quality, as listed in Table \ref{tab:image degradation techniques}.

\begin{table*}[t]
    \centering
    \caption{Image Manipulation Techniques used in our approach}
    \begin{tabular}{|p{5cm}|c|c|}
         \hline
         Operations & Parameters & Rationale \\
         \hline
         Random Crop, Vertical Crop,  \newline
         Horizontal Crop & {[0.4,0.6]} & Partial Subject\\
         \hline
         Color Jitter - Brightness,  \newline
         Contrast and Hue & {[0.3,0.6]} $\cup$ {[1.2,1.4]} & Poor Lighting \\
         \hline
         Gaussian Blur & {[0.8,1.2]} & Soft Images \\
         \hline
         Gaussian Noise & {[0.2,0.8]}  & Grainy Images \\
         \hline
         Grayscale & - & Image Filter\\
         \hline
         Random Rotation, & \{5,10,15,20\} & \\
         Random Rotation + Mixup & alpha - {[0.2,0.4]} & Camera Shake\\
         \hline

    \end{tabular}
    
    \label{tab:image degradation techniques}
\end{table*}

We have adopted a variety of techniques to generate synthetic training instances which emulate the low-quality images that we get in our product reviews, including (1) Vertical and horizontal crop - The location of the subject and object in an image do play an important role in defining the aesthetics of an image. Partial subject in an image do affect the aesthetics of an image, for instance, shirt buyers excludes the portion below the abdominal area while uploading pictures of them wearing the shirts they purchased. To mimic the same, good-quality images were subjected to cropping (Vertical and Horizontal). (2) Addition of color jittering by changing the Brightness, contrast and hue based on a scaling factor to mimic the poor lighting conditions. (3) Gaussian Blur and Gaussian noise to add fuzziness and graininess in an image. (4) Grayscale to mimic one simple basic filter which customers apply. Customers sometimes do use advanced filters, but we are not taking them into account for now. (5) Random Rotation and Rotation with Mixup are applied to achieve the camera shake effect. 

\subsection{Ranked Images Generation and Sampling Strategy}
In addition to what has already been described in section 3.1, It would be worthwhile to mention that UGC images were further divided into two classes, one in which the customer is actually wearing a product and the other being flat shot images. We achieved this using YOLOv5\cite{yolov5} model to classify images as with/without humans. We segregated them because in the case of apparel, images with usability have precedence and relevance when it comes to standalone images of products purchased.
With all the images, we make the following pairs - 
\[
\{(x,y): \exists x\in D_{+}, y \in D_{-}\}
\]

where $D_{+}$ and $D_{-}$ comes from the Table \ref{tab: image pair sampling strategy}.

\begin{table}[h]
    \caption {Image Pair Sampling Strategy}
    \begin{tabular}{c|c|c}
    \toprule
    S. No & $D_{+}$ & $D_{-}$ \\
    \hline
     1 &$D_{studio}$ & $D_{studio\_distorted}$ \\
     2 &$D_{ugc\_good}$   & $D_{ugc\_good\_distorted}$ \\
     3 &$D_{studio}$ & $D_{ugc\_good}$ \\
     4 &$D_{studio}$ & $D_{ugc\_bad}$ \\
     5 &$D_{ugc\_good\_human}$ & $D_{ugc\_bad\_human}$ \\
     6 &$D_{ugc\_good\_non\_human}$ &  $D_{ugc\_bad\_non\_human}$ \\
    \end{tabular}
    \label{tab: image pair sampling strategy}
\end{table}

We sample pairs as defined in the table uniformly. For pair 1, we considered studio images as positive samples and used image distortion techniques as described in section 3.2 in random order on positive samples to generate corresponding negative samples.
An analogous description holds for Pair 2. In pair 3,4 we considered studio images as positive samples and UGC images as negative samples. In pair 5, we considered "good" UGC images of users wearing the product as a positive sample and "bad" UGC images of users wearing the product as negative samples. In pair 6 we ranked "good" flat shot images higher than "bad" flat shot images.

\subsection{Neural Network Architecture}
\begin{figure}
    \centering
    \includegraphics[width=.45\textwidth]{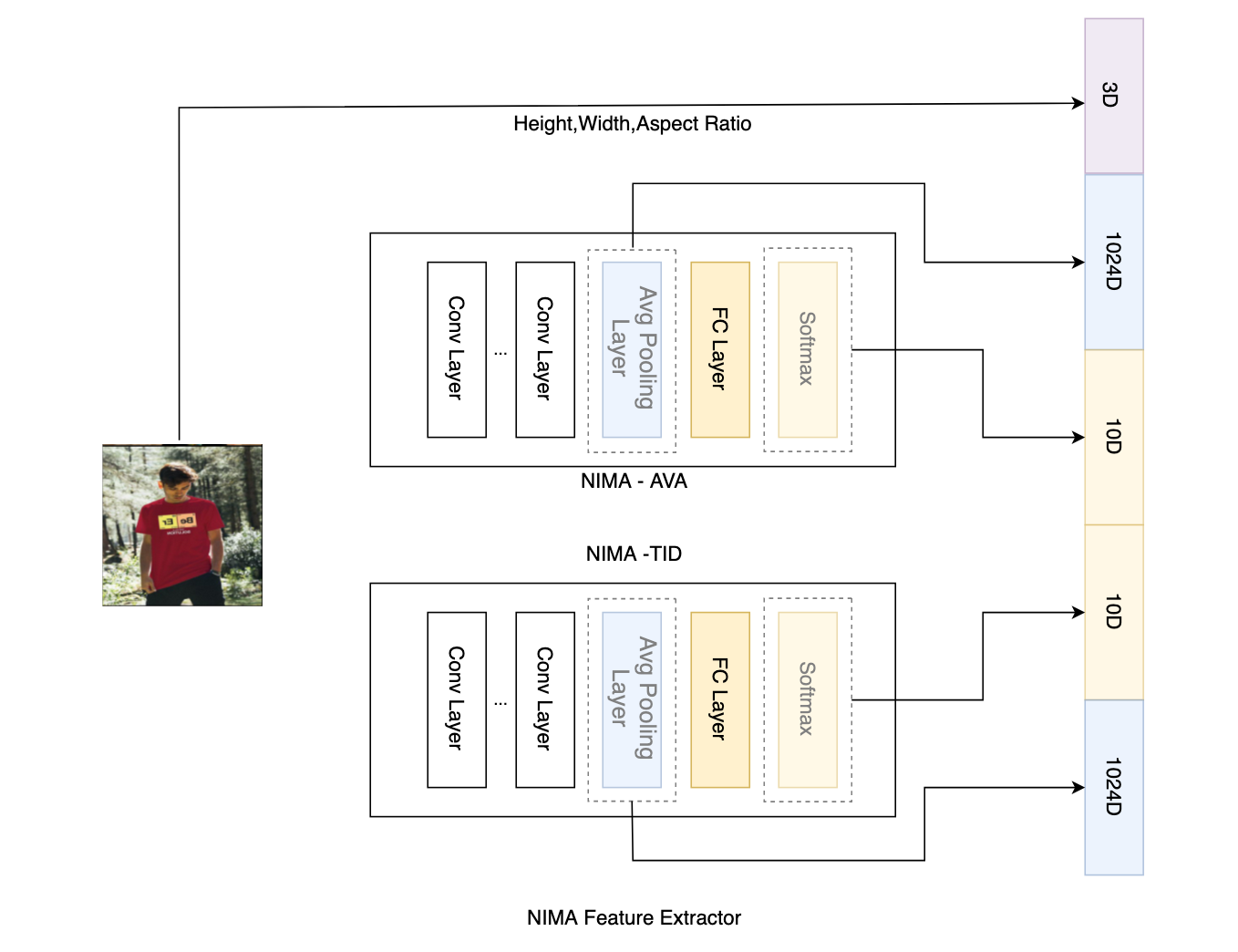}
    \caption{NIMA Aesthetic and Technical Model}
    \label{fig:NIMA}
\end{figure}

In our experiments, we have used NIMA\cite{talebi2018nima} as our reference to extract image features. As described in the NIMA paper, they have trained deep CNNs on two datasets. One network tries to capture the style, content, composition etc. and another network tries to capture the technical quality of an image. We aggregate the features generated from the penultimate layer, i.e. the average pooling layer, which is 1024 dimensions in our case. We have also aggregated the probability distribution of predicted rating as a feature, as described in Figure \ref{fig:NIMA}.

\begin{description}
    \item[NIMA - Aesthetics] This CNN model is based on MobileNet architecture whose weights are initialized by training on ImageNet dataset and then end-to-end training is performed on AVA\footnote{AVA images are obtained from \url{www.dpchallenge.com}, which is an online community for amateur photographers.} dataset. The AVA dataset contains 2,55,000 images, rated based on image aesthetics such as style, content, composition etc. by photographers. Each image is roughly rated by 200 people in response to a photography context on a scale of 1-10. This model tries to predict the normalized distribution of ratings for an image.

    \item[NIMA - Technical] This CNN model is based on MobileNet architecture. It is trained on the TID2013\cite{ponomarenko2015image} dataset. This contains 3000 images which are generated from 25 reference images, and 24 types of distortion with 5 levels of each distortion. Ratings are collected by showing a pair of distorted images for each reference image, and the observer has to select the better one in the pair. Unlike the AVA dataset, TID2013 provides just the mean opinion score and standard deviation. NIMA paper requires training on score probability distribution. The score distribution is approximated through maximum entropy optimization\cite{cover1999elements}. 
\end{description}

\begin{figure}
    \centering
    \includegraphics[width=.45\textwidth]{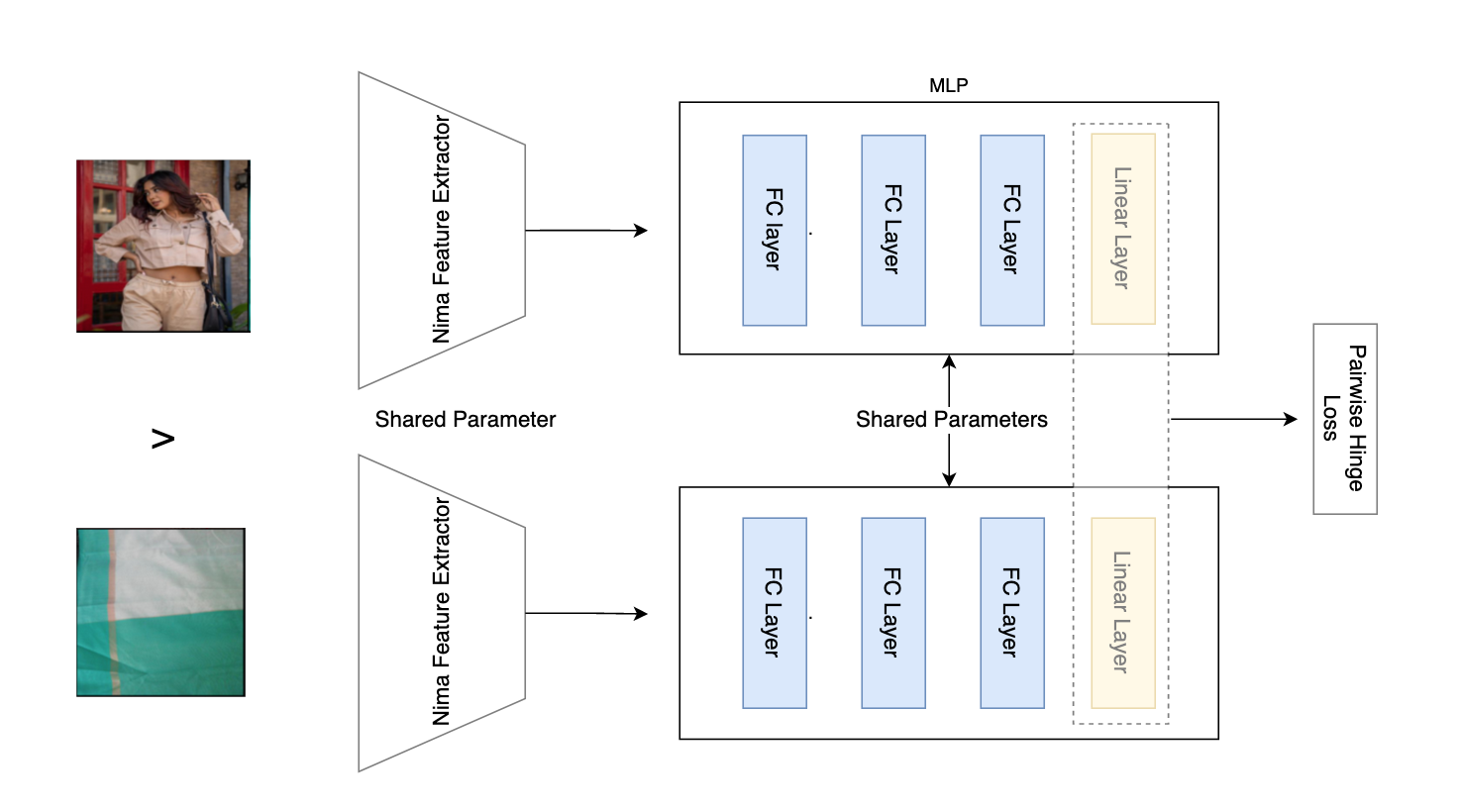}
    \caption{Network Architecture}
    \label{fig:overall-architecture}
\end{figure}

We have also taken the height, the width, and the aspect ratio of an image as features. The reason to incorporate them as features is that in NIMA we have to rescale all the images to a fixed size regardless of their original image aspect ratios. The lack of information about the original image size during the training of CNN in NIMA can affect its prediction, as the human rater may not give the same rating to the resized version of the image.

Given a pair of images $I_1$ and $I_2$ as input to the NIMA feature extractor, the output feature representation is denoted by $x_1$ and $x_2$ respectively. Now these features will be given as an input to our Siamese network shown in Figure \ref{fig:overall-architecture}. The output is represented by $f(x; \Theta)$ which is obtained by capturing the output of the last layer. Here $\Theta$ are the network parameters, here we will use $y$ to denote the ground truth value for the image. The network output of the final layer is a single scalar. The network is supposed to output higher scores for high-quality images and smaller scores for low-quality images. For a pair of images, the ranking loss is defined as -
\[
L_{rank} = \sum_{i,j} max(0,m-\delta(y_i \geq y_j)(f(x_i; \Theta) - f(x_j; \Theta)))
\]
where
\[\delta(y_i \geq y_j) = \begin{cases} 1, \text{ if } y_i \geq y_j\\ -1, \text{ if } y_i < y_j. \end{cases}
\]
where m is the minimal margin denoting the desired difference between the scores generated by the ranking network for a pair of images.
\section{Experiments}
\subsection{Implementation Details}
We train our Siamese network with the image pairs that we generated as described in section 3.3. We have set the hyperparameter m, which describes the minimal margin between the positive and negative image pair, to 1. 

We have collected 19K images from Myntra studio, 16K highly upvoted UGC images and 3.5K highly downvoted UGC images for generating training image pairs. 

For validation, we have kept 1000 images each from the studio, "good" UGC images and "bad" UGC images for 1000 different styles\footnote{At Myntra we use the term style to mean a specific product.}. Therefore, we have 3 images for each style. We have used accuracy (which is defined in section 5) as the metric on the validation set to select the best model.

We have used pre-trained NIMA feature extractor\cite{idealods2018imagequalityassessment} which is implemented in Keras\cite{chollet2015keras} and we have converted them into ONNX\cite{onnx}, as we do not train these networks. We run this ONNX model using ONNX Runtime\cite{ort}. 

Our MLP network contains 3 hidden layers and an output layer. The first hidden layer transforms the output from the NIMA feature extractor to 512 dimensions. Subsequent layers transform it to 256 and 128 dimensions, respectively. The network output of the final layer is a scalar value representing the score.

During training, input images are rescaled to 224 x 224. We train the network using ADAM optimizer. We have used the default learning rate($10^{-3}$) for fully connected layers and default weight decay regularization of $5*10^{-4}$. We have also used a learning rate scheduler, which will halve the learning rate if validation accuracy doesn't improve in five consecutive epochs. We have experimented with 16 as a batch size. All our implementation is done in PyTorch\cite{pytorch}.


\section{Results}
To compare our model, we use NIMA models as our baselines. Both NIMA-Aesthetics and NIMA-Technical predict the probability distribution of the score in the range 1-10, inclusive. We take the expected value of the NIMA-X's output on an image as the score of that image by model x. That is,

\[
f^X(I) = \sum_{i=1}^{i=10} i Pr_X(i; I) 
\]

where X is either A (for Aesthetics) or T (for Technical), and $Pr_X(i; I)$ denotes the probability for a score i by NIMA-X. Images with higher predicted scores are ranked higher.

\begin{figure}
    \centering
    \includegraphics[width=.5\textwidth]{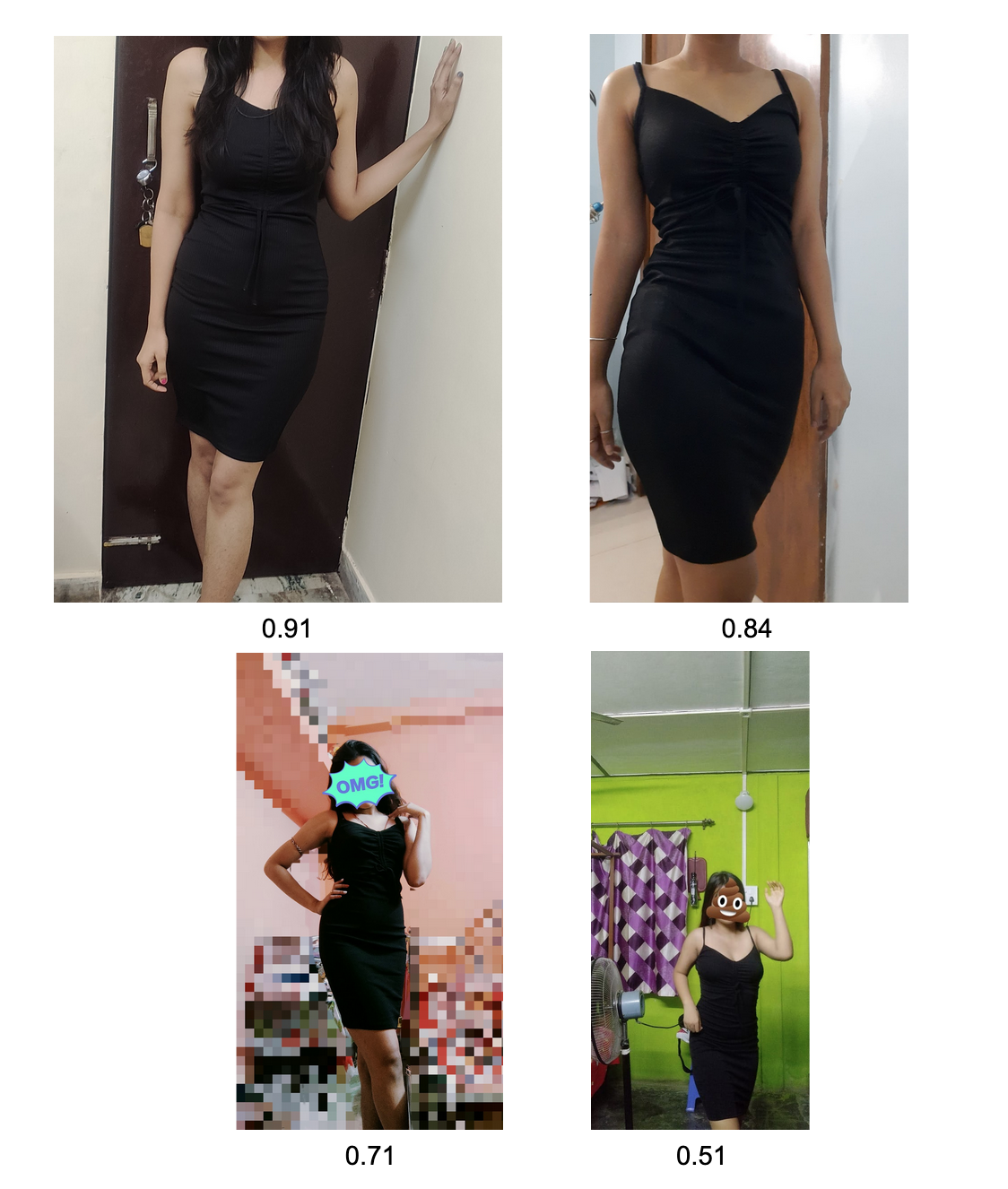}
    \caption{Sample test set images for a particular style. For each image, the number in the bottom center is the ground truth score for that image.}
    \label{fig:testset_sample}
\end{figure}

Since we do not have the ground truth rankings, we take users' ratings as a proxy for the quality of an image. That is, if a particular image $I$ has $u$ upvotes and $d$ downvotes, we assume that the ground truth quality score for that image is $\frac{u}{u+d}$. To create such a test set, we gathered around 850 images associated with highly engaged reviews from 20 popular styles. None of the images from these 20 styles (highly engaged or not) were kept in our training set or validation set. As mentioned in the introduction, the ratings are not given to the images, but the reviews. However, since these reviews are highly engaged, we assume that raters would have considered accompanying images while rating the review. To validate this hypothesis, we analyzed our reviews for their engagement and found that reviews with associated images had on average 6.5x engagement in terms of upvotes/downvotes as compared to reviews without images. We also did a paired t-test and found that this difference was statistically significant. A subset of the images with their computed scores is presented in Figure \ref{fig:testset_sample}.

We use two metrics to quantify our results. The first one is the Pearson correlation coefficient, which is a common metric to compare the performance of image ranking models on labeled datasets. It is computed as,

\[
    \rho(f, S) = \frac{\sum_{I \in S} (f(I) - \bar{f}(S))(g(I) - \bar{g}(S))}{\sqrt{\sum_{I \in S} (f(I) - \bar{f}(S))^2(g(I) - \bar{g}(S))^2}}
\]

where $f$ is either our baseline or our model, $g$ is the ground truth score function, $S$ is the set of images belonging to a particular style(i.e. all images in $S$ belong to the same product), $\bar{h}(S)$ is the mean of the scores of the images in $S$ computed by $h$. 0 correlation implies that the model gives scores that are unrelated to the ground truth scores (i.e. likes), positive correlation implies that the model scores highly liked images higher than highly disliked images. 


Another metric we use is accuracy. To compute accuracy for a particular style, we randomly (without replacement) pick 50 pairs of images for that style. We compute scores for these pairs using our model or the baselines. If we pick a pair $(I_1, I_2)$, and $g(I_1) > g(I_2)$, then models should generate $f(I_1) > f(I_2)$, otherwise that pair is considered as misclassified. 

Table \ref{tab:results} compares our approach with our baselines. We report the average of the metrics for all 20 styles. As can be seen in the first two rows of the table, NIMA models without finetuning, even though trained for aesthetics, outputs results that are completely uncorrelated with the proxy ground truth. Note that we did not train our model to predict the proxy scores. We are using a pairwise loss function to differentiate between positive and negative images, still, our model has a positive correlation with the proxy ground truth.  The accuracies of the NIMA models are no good than the accuracy one would get by guessing the binary prediction randomly.

\begin{table}[]
    \centering
    \caption{Quantitative results for our approach.}
    \begin{tabular}{|l|c|c|}
         \hline
         model & correlation coefficient & accuracy\\
         \hline
         NIMA-Aesthetics & -0.05 & 0.48\\
         NIMA-Technical & -0.02 & 0.50\\
         \hline
         Ours & \textbf{0.19} & \textbf{0.58} \\
         \hline
    \end{tabular}
    
    \label{tab:results}
\end{table}

\section{Conclusion and Future Work}
This paper presents an effective scheme of leveraging existing Deep learning models for Image Aesthetic Assessment as a feature extractor to fine-tune a Siamese network trained on synthetic data generated to rank UGC images.
We created the dataset by systematically degrading the quality of the studio and "good" UGC images. This was done to emulate the kind of low-quality imagery that we encounter on a routine basis.
We have seen that this approach helps in improving the ranking as compared to baseline NIMA-Aesthetics and NIMA-Technical. Our technique is not limited to NIMA feature extractor, as we can replace NIMA feature extractor with any other feature extractor e.g. \cite{ma2017lamp,sheng2018attention} trained on image aesthetics assessment datasets \cite{talebi2018nima, ponomarenko2015image}. 
Extending our existing approach to pre-train a CNN and fine-tune our labeled ranked dataset for fashion will be an interesting experiment to perform. We can leverage this model for Thumbnail generation, ranking catalog \& studio images. This model can be leveraged for providing customer feedback through a prompt about the quality of the image they have taken before they submit the images.


\bibliographystyle{ACM-Reference-Format}
\bibliography{sample-sigconf}

\end{document}